\documentclass[runningheads]{llncs}
\usepackage{amsmath} 
\usepackage[T1]{fontenc}
\usepackage{graphicx,verbatim}
\usepackage{subcaption}
\usepackage{amsfonts}
\usepackage[hidelinks]{hyperref}
\usepackage{color}
\usepackage[hidelinks]{hyperref}

\usepackage{booktabs} 
\usepackage{multirow}
\usepackage{xcolor}

\newcommand{\com}[1]{#1}

\newcommand{\subsd}[1]{_{\pm #1}}

\begin{document}
\title{SynSeq: End-to-End SYNTAX Score Prediction from Coronary Angiography Videos}
\titlerunning{SynSeq}
\author{Christoph Baumann\inst{1,2} \and 
Ronny Schweitzer\inst{3} \and 
Noemi Pavo\inst{3} \and 
Ulrike Attenberger\inst{4} \and 
Christian Loewe\inst{4} \and 
Philipp Seeböck\inst{1,2}} 
\authorrunning{Baumann et al.}

\institute{MANO Group, Computational Imaging Research Lab, Medical University of Vienna, Austria \and Comprehensive Center for Artificial Intelligence in Medicine, Medical University of Vienna, Austria \and Department of Cardiology, Medical University of Vienna, Austria \and Department of Biomedical Imaging and Image-Guided Therapy, Medical University of Vienna, Austria\\ \email{\{christoph.a.baumann,philipp.seeboeck\}@meduniwien.ac.at}}

\maketitle              
\begin{abstract}
The SYNTAX score is an established tool for assessing coronary artery disease and guiding revascularization treatment decisions. However, its manual estimation from coronary angiography videos by clinical experts is time-consuming and subject to inter-reader variability. While machine learning has shown promise in automating this process, prior work has primarily focused on lesion detection, characterization, or binary disease classification, leaving direct SYNTAX score prediction relatively unexplored. We propose \textit{SynSeq}, a video-based method for direct SYNTAX score prediction. It combines targeted preprocessing with a tailored training strategy using a zero-inflation-aware loss and linear target scaling. Evaluated on the public CardioSyntax dataset, SynSeq significantly outperforms previous state-of-the-art methods, improving $R^2$ by 0.55, reducing prediction bias by 93.1\% and achieving more consistent performance across annotations from three independent expert graders. In addition, SynSeq achieves a weighted $F_1$-score of 0.80 for revascularization treatment recommendations, slightly below inter-expert agreement. 
\com{These results demonstrate the potential of SynSeq to provide consistent, automated SYNTAX score assessment and reliable decision support for coronary revascularization planning.}


\end{abstract}

\section{Introduction}

Coronary artery disease (CAD) remains a leading cause of mortality worldwide, requiring precise diagnosis and management to optimize patient outcomes. When managing complex CAD, clinicians rely on the SYNTAX score as the established tool to grade lesion severity, complexity, and location from invasive coronary angiography (ICA) examinations \cite{serruys2009syntax}. Scores above a critical threshold ($\geq$33) strongly indicate superior long-term survival with coronary artery bypass grafting (CABG) over percutaneous coronary intervention (PCI) \cite{mohr2013coronary}. Recognizing its predictive power, the recent European Society of Cardiology (ESC) guidelines highlight that integrating machine learning into this workflow could significantly streamline the revascularization selection process and improve prognostic assessments \cite{vrints2024esc}. 
Despite its clinical utility, widespread adoption of the SYNTAX score remains bottlenecked. Manual calculation is limited by inter-observer variability due to anatomical ambiguities and is time-consuming, reducing its feasibility in everyday clinical practice \cite{garg2010syntax,genereux2011syntax}. In this work, we address this gap by introducing an automated, end-to-end framework capable of predicting patient-level SYNTAX scores directly from multi-view coronary angiography videos.

\paragraph{Related Work}
Prior automated coronary angiography efforts have focused on sub-tasks like vessel segmentation, lesion detection, and stenosis quantification rather than full SYNTAX score estimation \cite{Fawzi2026,lin2023stenunet,liu2023yolo,sun2026multisource,yang2026accurate}. While these methods have demonstrated promising performance on their respective tasks, they address only individual components of the reasoning required for SYNTAX score assessment. Moreover, the majority of prior work relies on single-frame datasets \cite{Mahmoudi2025,popov2024arcade}, limiting their ability to exploit the temporal and multi-view information inherent to angiography studies.
Recent methods have leveraged full video sequences to exploit temporal and multi-view projections. CathAI~\cite{avram2023cathai} introduces an end-to-end pipeline for automated ICA interpretation and stenosis estimation. DeepCoro~\cite{Labrecque2024Coro} and DeepCORO-CLIP~\cite{harrabi2025deepcoro} further advanced video-based representation learning through large-scale spatiotemporal and multi-view pretraining. However, these still focus primarily on stenosis localization and anatomical characterization rather than direct SYNTAX score estimation, which requires integrating lesion severity, vessel dominance, and anatomical complexity. 
The CardioSyntax benchmark \cite{ponomarchuk2025cardiosyntax} is the first large-scale dataset for end-to-end SYNTAX score prediction from multi-view videos. While the accompanying baseline models include continuous score prediction, they primarily emphasize zero-versus-nonzero discrimination (healthy versus diseased). While useful for screening, this provides limited insight into high-complexity cases where clinical decision-making between PCI and CABG is most critical. Furthermore, the zero-inflation and long-tailed distribution properties of the SYNTAX score task remain unaddressed, limiting performance on highly diseased, underrepresented cases. Our work directly targets these challenges, focusing on robust continuous score prediction across the full range of SYNTAX values.

\paragraph{Contribution} We introduce \textit{SynSeq}, a video-based framework for automated SYNTAX score prediction that focuses on directly optimizing the continuous scoring objective. \com{We build SynSeq on the previously introduced Multi-View-Look (MVL) method~\cite{ponomarchuk2025cardiosyntax}}. Our main contributions are: (i) We propose targeted preprocessing for multi-view coronary angiography videos to enhance feature extraction from complex temporal and cross-view inputs. (ii) We design a tailored optimization strategy with two distinct adjustments to handle the highly skewed data distribution: (1) a sample-weighting penalty function to counteract skewed label distribution, and (2) a linear rather than logarithmic target scaling to preserve score granularity in high-severity cases. (iii) We present a rigorous evaluation on the full CardioSyntax dataset, demonstrating significantly improved performance over prior state-of-the-art methods and increased robustness across multiple expert annotations.

\section{Method}
\begin{figure}[t]
    \centering
    \includegraphics[width=1\linewidth]{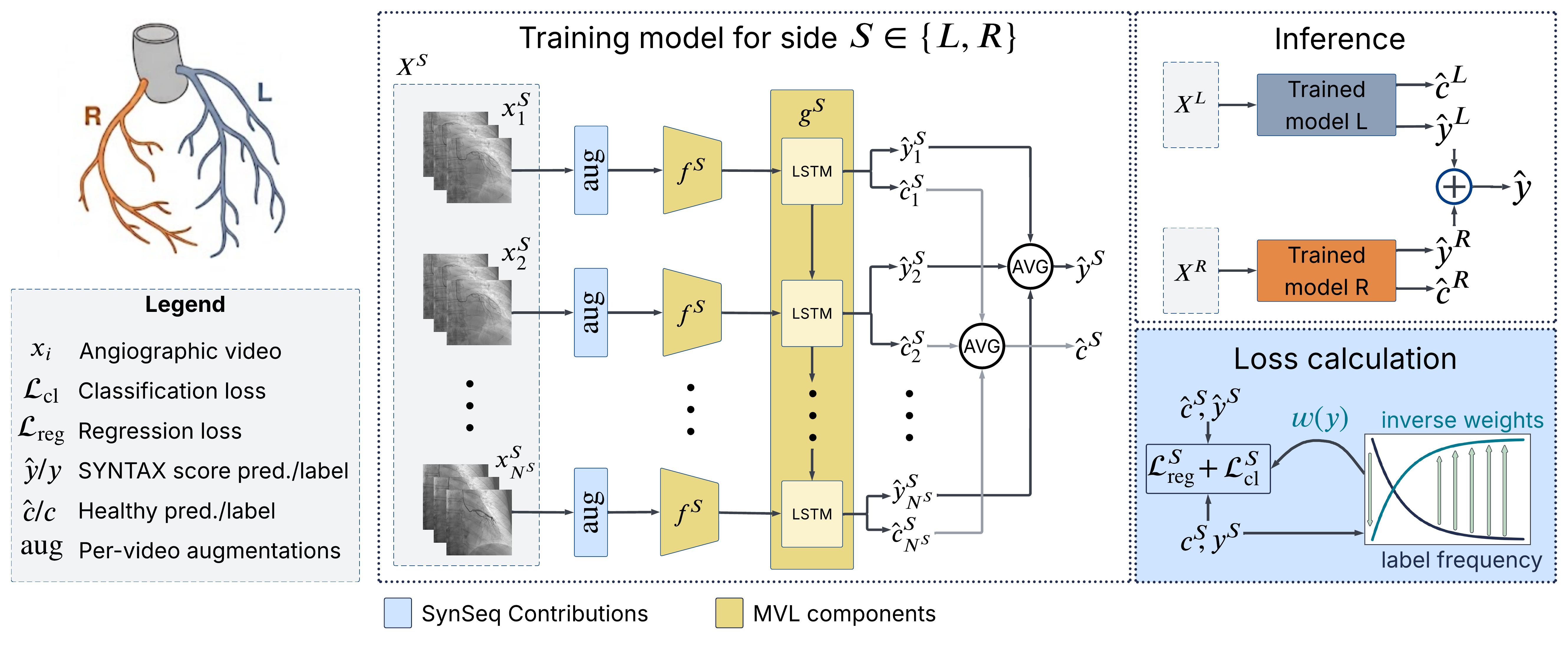}
    \caption{Overview of the proposed \textit{SynSeq} method. Coronary angiographic video sequences of the left ($X^L$) and right ($X^R$) vessel tree are processed by two independent branches with identical architecture but separate weights. Each sequence is encoded by a 3D-ResNet ($f^S$), with resulting features being fed into a LSTM-based dual-head predictor ($g^S$), estimating a SYNTAX score ($\hat{y}^S$) and vessel status ($\hat{c}^S$) per branch. Inverse frequency weighted losses ($\mathcal{L}_{\mathrm{reg}}^S$, $\mathcal{L}_{\mathrm{cl}}^S$) are computed independently for each branch. During inference, summing left and right predictions results in a patient-level SYNTAX score $\hat{y} = \hat{y}^L + \hat{y}^R$. The temporary head $h$ used in pre-training is omitted.
    }
    \label{fig:method_figure}
\end{figure}

Our method \textit{SynSeq} predicts the patient-level \textit{\textbf{SYN}}TAX score $\hat{y} \in \mathbb{R}$ from an arbitrary number of multi-view angiographic video \textit{\textbf{seq}}uences $X$, using a backbone network $f$ and prediction head $g$ (Figure \ref{fig:method_figure}). For each patient, SynSeq processes left and right coronary artery trees independently, with the corresponding sets of angiography video sequences denoted as $X^S$ for $S \in {L, R}$. The left and right sets are defined as $X^L = \{x^L_i\}^{N^L}_{i=1} $ and $X^R = \{x^R_i\}^{N^R}_{i=1}$, where $N^L$ and $N^R$ denote the number of video sequences per side. Each set contains at least one video and is treated as an unordered collection. Each coronary artery tree is processed by an independent 3D-ResNet backbone $f^S$ with parameters $\theta_f^S$ to extract per-video feature representations, followed by a dual-head LSTM $g^S$ with parameters $\theta_g^S$ to obtain predictions for each video. The LSTM outputs (i) a scalar estimate of the SYNTAX sub-score and (ii) a binary logit indicating the probability of disease presence (SYNTAX score $>0$). To account for varying numbers of acquisitions, we obtain $(\hat{y}_i^S, \hat{c}_i^S) = g^S(f^S(x_i^S))$ and compute the side-specific predictions by averaging:
\begin{equation}
(\hat{y}^S, \hat{c}^S) = \frac{1}{N^S} \sum_{i=1}^{N^S} (\hat{y}_i^S, \hat{c}_i^S).
\end{equation}

During inference, the final patient-level SYNTAX score is obtained by summing contributions from both coronary trees, with negative values clamped to zero:
\begin{equation}
    \hat{y} = \mathrm{max}(0, \hat{y}^L) + \mathrm{max}(0,\hat{y}^R)
    \label{eq:zeroclamp}
\end{equation}

\paragraph{Inputs \& Preprocessing} 
To simulate realistic acquisition artifacts, domain tailored data augmentation is applied during training. Random spatial cropping is performed with a relative crop area $\kappa \in [\kappa_{\min}, \kappa_{\max}]$ and an aspect ratio $\eta\in[\eta_{\min}, \eta_{\max}]$. Minor video rotations are sampled from $\rho \in [-\rho_{\max}, \rho_{\max}]$, and brightness and contrast are adjusted by a multiplicative factor $\delta_{\mathrm{int}}$. The backbone input is reduced to a single channel for grayscale angiography, using training-set mean $\mu$ and standard deviation $\sigma$ for input normalization. Unlike standard three-channel color image augmentations as used in \cite{ponomarchuk2025cardiosyntax}, these adjustments are tailored to X-ray angiography to mimic real-life acquisition variance. Our method assumes each patient has at least one video for both the left and right coronary vessel trees,  labeled by artery side and grouped by patient ID. For all videos, a temporal center crop of length $\tau$ is used; sequences shorter than $\tau$ are padded via temporal repetition.

\paragraph{Continuous long-tailed distribution handling} 
We explicitly address the zero-inflated and skewed distribution of SYNTAX scores, characterized by a high prevalence of healthy cases ($y_i^S = 0$) and a sparse distribution of diseased samples ($y_i^S \geq 1$). Specifically, each sample is assigned a weight inversely proportional to the frequency of its ground-truth sub-score $y_i^S$:

\begin{equation}
w(y_i^S) = \left( \sum_{j=1}^{T} \mathbb{I}\left( \lfloor y_j^S \rfloor = \lfloor y_i^S \rfloor \right) \right)^{-1}
\end{equation}
where the function $\mathbb{I}$ is the identity function that maps the boolean input to 0 or 1 and T denotes the total number of training samples of all patients for a given side $S$.

\paragraph{Loss functions} SynSeq is trained using a multi-task objective combining regression and binary classification. The regression head uses a sample-weighted Mean Squared Error (MSE) loss, with targets scaled to $[0,1]$ via a normalization parameter $\lambda$ that is utilized instead of logarithmic scaling used in \cite{ponomarchuk2025cardiosyntax}
\begin{equation}
\mathcal{L}_{\mathrm{reg}}^{S}(\hat{y}^{S}, y^{S}) = \frac{1}{M} \sum_{k=1}^{M} w(y_k^S) \cdot \left( \hat{y}^{S}_{k} - \frac{y^{S}_{k}}{\lambda} \right)^2
\end{equation}
where $M$ denotes the total number of samples in the respective mini-batch.
In addition,  the classification heads predict whether a system is diseased ($c_k^S = 1$ if $y_k^S > 0$, else $0$) via a sample-weighted Binary Cross-Entropy loss:

\begin{equation}
\mathcal{L}_{\mathrm{cl}}^{S}(\hat{c}^S,c^S) = -\frac{1}{M} \sum_{k=1}^M w(y_k^S) \cdot \left[ c_k^S \log \sigma(\hat{c}_k^S) + (1 - c_k^S) \log (1 - \sigma(\hat{c}_k^S)) \right]
\end{equation}
The final unified multi-task loss for each coronary artery tree ($S \in \{L,R\}$) combines both losses, balanced by hyperparameter $\alpha$:
\begin{equation}
\mathcal{L}^L = \mathcal{L}^L_{\mathrm{reg}} + \alpha \mathcal{L}^L_{\mathrm{cl}},\quad \mathcal{L}^R = \mathcal{L}^R_{\mathrm{reg}} + \alpha \mathcal{L}^R_{\mathrm{cl}}
\end{equation}

\paragraph{Training procedure} For each coronary arterial tree $L$ and $R$, networks $f^S$ and $g^S$ are trained in separate optimization loops. Parameters $\theta_f^L$, $\theta_g^L$ and $\theta_f^R$, $\theta_g^R$ are learned using the corresponding left- and right-branch losses. In both cases, training follows a two-stage procedure, conducted independently for each tree. First, the backbone network $f^S$ is pretrained on a per-video classification task. A temporary linear head $h^S$ maps the backbone features to a single logit predicting the health probability for each side $\hat{c}^S_{\mathrm{pre}} = h^S\left( f^S(x_i^S) \right)$ and is trained with the binary cross entropy-loss $\mathcal{L}_{\mathrm{cl}}^{S}(\hat{c}^S_{\mathrm{pre}},c^S)$.
Second, $h^S$ is discarded and the full architecture is trained by optimizing the dual-head LSTM $g^S$ while the backbone $f^S$ is frozen, followed by a joint fine-tuning of both. The sub-scores are aggregated only during inference to yield the final patient-level SYNTAX score. \com{This overall training procedure recreates the training steps from the MVL method~\cite{ponomarchuk2025cardiosyntax}. }The implementation will be released at \url{https://github.com/cirmuw/SynSeq}.

\section{Experimental Setup}

\paragraph{Data} We use the publicly available CardioSyntax dataset~\cite{ponomarchuk2025cardiosyntax} for our experiments. The dataset comprises 1,844 patients, providing a total of $9,590$ left coronary artery (LCA) and $3,970$ right coronary artery (RCA) angiographic videos. Each video has a resolution of $512\times512$ pixels at 15 frames per second, with lengths ranging from 2 to 355 frames (90\% between 26 and 77). The ground-truth SYNTAX sub-scores are provided for the LCA and RCA separately by a single primary interventional cardiologist. For a subset of 60 patients, consensus labels from two additional independent interventional cardiologist experts are provided within the dataset repository.

\paragraph{Experiments}
\com{We evaluate SynSeq against the Multi-View-Look (MVL) baseline~\cite{ponomarchuk2025cardiosyntax}, a baseline based on the domain-pretrained DeepRV model~\cite{Fawzi2026} that utilizes a single view per patient and is not fine-tuned beyond linear probing in our experiments, and the naive mean baseline for better context.} To ensure reproducibility, we establish a strict patient-wise split: $60$ multi-expert annotated patients are held out as an independent test set, and the remaining data is partitioned via 5-fold cross-validation. Only for final performance evaluation, models from each fold are evaluated on the test set, with performance reported as mean $\pm$ standard deviation across the folds.

We assess regression performance using the coefficient of determination ($R^2$), mean absolute error (MAE), and mean error (bias). \com{$R^2$ is also provided for inter-expert agreement.} To determine if our method significantly reduces MAE and bias magnitude compared to baselines, a one-tailed Wilcoxon signed-rank test is performed on patient-level ensemble predictions, obtained by averaging outputs from the five-fold models. Particularly this is done for the sample-level metrics (MAE, bias), while $R^2$ is excluded from significance testing as it is an aggregate dataset-level metric.  Additionally, we evaluate the revascularization treatment recommendation performance using the weighted $F_1$-score computed over the standard risk categories: low ($\leq$22), intermediate (23–32), and high ($\geq$33) risk. We report the (macro) precision, recall, F1-score and accuracy as mean $\pm$ standard deviation across the folds. \com{A boundary analysis of misclassified patients describes the deviance in terms of distance between the predicted SYNTAX score and the correct class. Furthermore, we provide a Bland-Altman analysis to investigate systematic errors.}
\com{We also report the binary classification performance of the classification head after pretraining and in the final model.} 

\com{To assess the sensitivity of our performance estimates to data partitioning, we evaluate two partition schemes: The 60-patient consensus test split (Split I) and an 80/20 SYNTAX-stratified split of the full cohort with these patients in the evaluation set (Split II). Results use Split I unless stated otherwise.}

\paragraph{Ablations} To isolate design effects, we evaluate individual components by cumulatively removing them from the proposed method and measuring changes in $R^2$, MAE, and bias. Specifically, we analyze: (i) the impact of long-tailed distribution handling via sample-weighting penalty removal in the loss (\textit{SW}); (ii) the effect of optimizing in log-space instead of linear target space (\textit{LinS}); and (iii) the contribution of domain-specific angiography video preprocessing (\textit{APre}).

\paragraph{Implementation Details}
SynSeq uses a Kinetics-400 pretrained 3D-ResNet18 (R3D-18) backbone $f$~\cite{Kaye2017Kinetics,Tran2017resnet} mapping videos to 512-D embeddings, and a single-layer projected LSTM $g$ (512 input, 128 hidden) directly outputting $\hat{y}^{S}$ and $\hat{c}^{S}$. All stages use Adam (batch size 4, 30 epochs) and a OneCycle learning rate (LR) scheduler. Training follows two phases: (1) pretraining $f$ (max LR $10^{-4}$), and (2) training $g$ with $f$ frozen (max LR $10^{-4}$), followed by joint end-to-end fine-tuning with $f$ unfrozen (max LR $10^{-5}$). \com{Here, we followed the hyper-parameter configuration of the MVL method~\cite{ponomarchuk2025cardiosyntax}.}
Other parameters include augmentation settings ($\kappa_{\min}=0.9$, $\kappa_{\max}=1$, $\eta_{\min}=0.9$, $\eta_{\max}=1.1$, $\rho_{\max}=5^\circ$, $\delta_{\mathrm{int}}=0.1$), normalization constants ($\mu=0.55871$, $\sigma=0.151$), loss weight $\alpha=1.0$, crop length $\tau=32$, and scale factor $\lambda=70$.

\section{Results}
\begin{figure}[t]

    \centering
    \includegraphics[width=\linewidth]{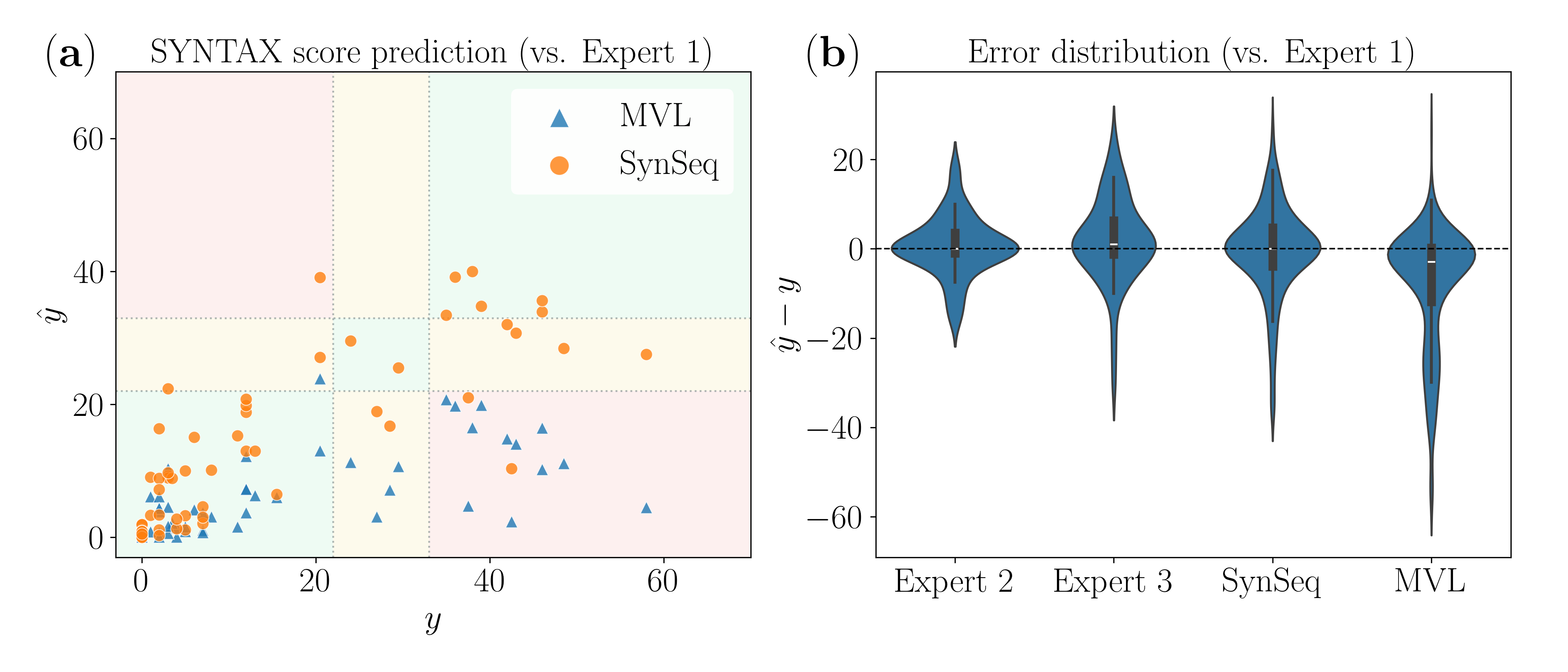}

    \caption{Performance comparison with expert 1 as reference. (a) Scatter plot of predicted $\hat{y}$ vs expert 1 SYNTAX scores $y$, with revascularization zones (green: agreement; yellow/red: minor/major disagreement). (b) Violin plot of SYNTAX errors $\hat{y}-y$ for MVL, SynSeq, and experts 2\&3 relative to expert 1.} 
    \label{fig:expert_comparison}
\end{figure}

SynSeq clearly outperforms the baselines across all metrics and evaluation scenarios. \com{Table \ref{tab:main_results} shows regression performance, with SynSeq achieving an $R^2$ of $0.62$ compared to $0.07$ for MVL~\cite{ponomarchuk2025cardiosyntax} and $-0.11$ for DeepRV~\cite{Fawzi2026}. The gap widens on diseased samples (SYNTAX score $>0$), with $R^2$ $0.54$ vs. $-0.14$/$-0.17$, bias $-0.90$ vs. $-10.47$/$-2.28$, and MAE $8.24$ vs. $11.90$/$14.09$.} The improvement is consistent across folds and statistically significant (bias: p<0.001, MAE: p<0.05). \com{This demonstrates reduced systematic bias of SynSeq, with substantially less underestimation of SYNTAX score compared to both MVL and DeepRV, which is also reflected in the scatter plot in Fig.~\ref{fig:expert_comparison}(a). Split II with the extended evaluation set gives the same picture (SynSeq/MVL:  $R^2$ 0.61/0.15, bias -1.21/-8.03 , MAE 6.82/9.13), indicating stability of performance estimates with respect to data partitioning.}
Fig.~\ref{fig:expert_comparison}(a) further illustrates revascularization treatment recommendations performance for SynSeq and MVL, with decision boundaries separating the low-, intermediate- and high-risk categories. SynSeq improves classification performance over the MVL baseline (weighted $F_1$: $0.80\pm0.04$ vs. $0.64\pm0.05$), with more predictions falling into the correct risk category (Fig.~\ref{fig:expert_comparison}(a)). \com{Its performance is slightly below inter-expert agreement, with weighted $F_1$-scores of 0.85 between experts 1 and 2, and 0.82 between experts 1 and 3.}

\com{SynSeq misclassified 11 of 60 patients across the clinical SYNTAX cutoffs ($\leq22$, $23$-$32$, $\geq33$), compared to 17 of 60 for MVL. Its errors were both smaller (4.2 vs. 9.8 distance in SYNTAX points from the true class boundary on average, over misclassified patients) and less consequential: only 3 crossed multiple decision thresholds in a way that would alter treatment strategy, versus 12 for MVL. Bland–Altman analysis supports this with a lower mean bias (0.57 vs. 8.29) and narrower 95\% limits of agreement ([-17.76, 18.89] vs. [-17.34, 33.92]).}

\begin{table}[t]
\centering
\caption{Regression performance on the test set using Expert 1 labels. $\bar{y}$ denotes the naive mean baseline. Bold: best per column.}
\begin{tabular}{l ccc ccc}
\toprule
& \multicolumn{3}{c}{\textbf{All samples}} & \multicolumn{3}{c}{\textbf{Only SYNTAX $>$ 0}} \\
\cmidrule(r){2-4} \cmidrule(l){5-7}
\textbf{Method} & $R^2$ $\uparrow$ & bias & MAE $\downarrow$ & $R^2$ $\uparrow$ & bias & MAE $\downarrow$ \\
\midrule
$\bar{y}$  & $-0.18$ & $-6.92$ & $12.23$ & $-0.40$ & $-10.40$ & $13.53$ \\
\com{DeepRV~\cite{Fawzi2026}} & $-0.11\subsd{0.11}$ & $0.78\subsd{3.23}$ & $13.88\subsd{1.49}$  & $-0.17\subsd{0.1}$ & $-2.28\subsd{3.16}$ & $14.09\subsd{1.07}$ \\
MVL~\cite{ponomarchuk2025cardiosyntax} & $\phantom{-}0.07\subsd{0.07}$ & $-8.29\subsd{0.63}$ & $9.62\subsd{0.59}$  & $-0.14\subsd{0.08}$ & $-10.47\subsd{0.80}$ & $11.90\subsd{0.72}$ \\
SynSeq         & $\mathbf{\phantom{-}0.62 \subsd{0.06}}$ & $\mathbf{-0.57\subsd{1.72}}$ & $\mathbf{6.74\subsd{0.29}}$ & $\mathbf{0.54\subsd{0.07}}$ & $\mathbf{-0.90\subsd{2.14}}$ & $\mathbf{8.24\subsd{0.45}}$ \\
\bottomrule
\end{tabular}
\label{tab:main_results}
\end{table}

\begin{table}[t]
\centering
\caption{Model performance in $R^2$ across three independent experts scores.}

\begin{tabular}{lccc}
\toprule
 & \textbf{Exp. 1} & \textbf{Exp. 2} & \textbf{Exp. 3} \\
\midrule
$\bar{y}$     & -0.18 & -0.21 & -0.33 \\
\com{DeepRV~\cite{Fawzi2026}} & $-0.11\subsd{0.11}$& $-0.03\subsd{0.12}$ & $-0.07\subsd{0.13}$ \\
MVL~\cite{ponomarchuk2025cardiosyntax} & $\phantom{-}0.07\subsd{0.07}$ & $\phantom{-}0.03\subsd{0.07}$ & $-0.10\subsd{0.04}$ \\
SynSeq & $\mathbf{\phantom{-}0.62\subsd{0.06}}$ & $\mathbf{\phantom{-}0.60\subsd{0.06}}$ & $\mathbf{\phantom{-}0.60\subsd{0.10}}$ \\
\bottomrule
\label{tab:expert_table}
\end{tabular}
\end{table}
\begin{table}[t]

    \centering
    \caption{\com{Ablation analysis of SynSeq components evaluated against Expert 1.}}
    \label{tab:ablations}
    \begin{tabular}{l ccc ccc}
    \toprule
    \textbf{Method} & SW & LinS & APre & $R^2$ $\uparrow$ & bias & MAE $\downarrow$ \\
    \midrule
    SynSeq         & \checkmark & \checkmark & \checkmark & $\phantom{-}0.62\subsd{0.06}$ & $-0.57\subsd{1.72}$ & $6.74\subsd{0.29}$ \\
    \cmidrule(lr){1-4}
    SynSeq$_{LinS-APre}$ & X & \checkmark & \checkmark & $\phantom{-}0.50\subsd{0.04}$ & $-4.58\subsd{0.27}$ & $7.27\subsd{0.23}$ \\
    \com{SynSeq$_{LinS}$}   & X & \checkmark & X & $\phantom{-}0.49\subsd{0.05}$ & $-1.81\subsd{1.31}$ & $7.84\subsd{0.49}$ \\
    \com{SynSeq$_{APre}$}   & X & X & \checkmark & $\phantom{-}0.38\subsd{0.05}$ & $-6.59\subsd{0.48}$ & $7.84\subsd{0.29}$ \\
    \com{SynSeq$_{SW}$}  & \checkmark & X & X  & $\phantom{-}0.07\subsd{0.34}$ & $-7.09\subsd{3.68}$ & $9.58\subsd{1.84}$ \\
    \bottomrule
    \end{tabular}
\end{table}

Table~\ref{tab:expert_table} summarizes performance across all of the three independent experts. SynSeq achieves consistent $R^2$ of 0.60 to 0.62 across all experts, indicating robustness to inter-rater variability. Although these values are below the highest inter-expert agreement ($R^2=0.83$ between experts 1 and 2), they are close to the agreement observed between experts 1 and 3 ($R^2=0.65$) and experts 2 and 3 ($R^2=0.67$). Figure~\ref{fig:expert_comparison}(b) supports these findings by showing reduced error variability for SynSeq relative to the MVL baseline, approaching the variability observed between human experts.

Table~\ref{tab:ablations} shows the SynSeq ablation results, demonstrating the relevance of all proposed components. Removing sample-weighting in the loss (\textit{SW}) substantially increases systematic bias while degrading MAE and $R^2$. \com{This drop is also reflected in risk category classification performance, with weighted F1-score reduction from 0.8 to 0.68.} Replacing the linear optimization objective with a log-transformed objective (\textit{LinS}) also increases bias and reduces $R^2$. Removing the preprocessing adaptations (\textit{APre}) leads to a substantial deterioration across all evaluation metrics as well. 
\com{Zero-clamping introduced in Eq. \ref{eq:zeroclamp} changes the result slightly, with an average of 16.6 samples out of the 60 test samples being clamped for each side across folds. If zero-clamping is not applied, $R^2$ decreases to $0.58\pm0.07$, bias increases to $-2.15\pm1.23$ and MAE increases to $7.87\pm1.17$.}

\com{
The separate binary classification head (healthy vs. diseased) achieved a macro F1-score of $0.727 \pm 0.008$ 
and an accuracy of $0.735 \pm 0.008$ after pretraining, increasing to $0.747 \pm 0.047$ macro F1 and $0.783 \pm 0.051$ accuracy after full training. The MVL baseline shows the same trend, but remains below these values ($0.703 \pm 0.020$  macro F1, $0.743 \pm 0.023$ accuracy).}

\section{Discussion and Conclusion}
SynSeq demonstrates strong agreement with expert-derived SYNTAX scores and achieves performance approaching inter-expert variability for clinically relevant revascularization decisions. The model maintains stable performance across three independent expert annotators despite being trained using labels from a single expert. This suggests that SynSeq learns a robust representation that is aligned with underlying angiographic patterns rather than overfitting to individual annotation styles.
Compared to the MVL baseline~\cite{ponomarchuk2025cardiosyntax}, SynSeq consistently reduces error and improves agreement with expert annotations. A key observation is that model improvements emerge from the interaction of architecture design, target formulation, and domain-specific preprocessing. First, optimizing in log-space was inferior to linear-space training, which is consistent with the additive structure of the SYNTAX score. Since the SYNTAX score is constructed as a cumulative severity index, linear-space regression better preserves proportional differences between cases, whereas log-space optimization may compress clinically relevant separations. Second, explicit sample-weighting reduces systematic underestimation, which is particularly important in clinical decision thresholds where small score shifts can alter CABG versus PCI recommendations. Third, coronary angiography videos exhibit substantial variability due to projection angles, motion, contrast dispersion, and field-of-view differences. We hypothesize that domain-specific preprocessing and augmentation leads to a more consistent representation of vessel structures across frames and promotes invariance to task-irrelevant factors. 

\paragraph{Limitations} \com{CardioSyntax lacks an official split, limiting direct comparability with its reported results \cite{ponomarchuk2025cardiosyntax}. The original study reports 5-fold cross-validation without a separate held-out test set, thus the evaluation protocol may differ from ours. Moreover, while we used a patient-level split, we cannot determine whether the original evaluation did as well. The reported metrics therefore seem to estimate different quantities, so we do not read them as directly comparable. To address this, we perform cross-validation across multiple seeds, define a held-out test cohort scored by all three experts, and confirm stability of performance estimates across data partitions.}

\com{Another limitation is the small number of baselines, as model weights and training data are not publicly available for CathAI~\cite{avram2023cathai}, DeepCoro~\cite{Labrecque2024Coro} and DeepCoroCLIP~\cite{harrabi2025deepcoro}.}
Natural clinical prevalence causes class imbalance affecting model calibration and performance, and our results show that applying explicit mitigation techniques is essential for automated SYNTAX score prediction.
Finally, although SynSeq shows robust performance in the public CardioSyntax dataset, external validation across institutions and imaging protocols is required to assess generalizability beyond the current dataset. 

\paragraph{Conclusion}
This work presents SynSeq, a method for predicting SYNTAX scores directly from multi-view coronary angiography. The results indicate that accurate and stable prediction requires (i) domain-specific angiography preprocessing to reduce acquisition variability and enforce invariance to task-irrelevant factors, (ii) optimization in a linear target space consistent with the additive structure of SYNTAX scoring, and (iii) explicit handling of long-tailed score distributions to reduce systematic bias. \com{Together, these design choices yield significant performance improvements compared to state-of-the-art and remains stable across independent annotators.} SynSeq demonstrates the potential of learning-based SYNTAX estimation to support more consistent and scalable cardiovascular treatment planning.

\begin{credits}
\subsubsection{\ackname}

This research was conducted within an Inter-University Cluster Project jointly funded by the University of Vienna and the Medical University of Vienna. AI-POD has received funding from the European Union’s Horizon Europe research and innovation programme under grant agreement 101080302, and from the Swiss State Secretariat for Education, Research and Innovation (SERI) under contract number 23.00174. Views and opinions expressed are however those of the author(s) only and do not necessarily reflect those of the European Union or HaDEA. Neither the European Union nor the granting authority can be held responsible for them.

\end{credits}

%
%
%

\bibliographystyle{splncs04}
\bibliography{mybibliography}
\end{document}